\RequirePackage{amsmath}
\documentclass[runningheads]{llncs}
\usepackage[T1]{fontenc}
\usepackage{graphicx}
\usepackage{amssymb}
\usepackage[short]{optidef}

\usepackage{hyperref}

\usepackage{booktabs}
\usepackage{csquotes}
\usepackage[table]{xcolor}

\usepackage{cleveref}
\usepackage[acronym]{glossaries}
\usepackage{glossaries-extra}

\usepackage{caption}

\newcommand{\nR}{\mathbb{R}}
\newcommand{\dif}[1]{\mathrm{d}{#1}\,}
\newcommand{\CK}{\text{CK}}

\setabbreviationstyle[acronym]{long-short}
\newacronym{ai}{AI}{Artificial Intelligence}
\newacronym{api}{API}{Application Programming Interface}
\newacronym{ml}{ML}{Machine Learning}
\newacronym{ann}{ANN}{Artificial Neural Network}
\newacronym{pp}{PP}{Piecewise Polynomial}
\newacronym{pinn}{PINN}{Physics-informed Neural Network}
\newacronym{rl}{RL}{Reinforcement Learning}

\begin{document}
\title{Energy Optimized Piecewise Polynomial Approximation Utilizing Modern Machine Learning Optimizers}
\titlerunning{Energy Optimized Piecewise Polynomial Approximation}
%
\author{Hannes Waclawek\inst{1, 2}\orcidID{0000-0003-3715-6845} \and
Stefan Huber\inst{1}\orcidID{0000-0002-8871-5814}}
\authorrunning{H. Waclawek \and S. Huber}
%
\institute{Josef Ressel Centre for Intelligent and Secure Industrial
Automation,\\Salzburg University of Applied Sciences, Austria \and Paris Lodron
University Salzburg, Austria\\
\email{\{hannes.waclawek, stefan.huber\}@fh-salzburg.ac.at}}
\maketitle              
\begin{abstract}
    This work explores an extension of \gls{ml}-optimized \gls{pp}
    approximation by incorporating energy optimization as an additional
    objective. Traditional closed-form solutions enable continuity and
    approximation targets but lack flexibility in accommodating complex
    optimization goals. By leveraging modern gradient descent optimizers within
    TensorFlow, we introduce a framework that minimizes elastic strain energy
    in cam profiles, leading to smoother motion. Experimental results confirm
    the effectiveness of this approach, demonstrating its potential to
    Pareto-efficiently trade approximation quality against energy consumption.
\keywords{Piecewise Polynomials \and Gradient Descent \and Approximation 
          \and TensorFlow \and Electronic Cams \and Energy Optimization}
\end{abstract}
\section{Introduction}

\subsection{Motivation and Problem Statement}

\glspl{pp} are commonly utilized in various scientific and engineering
disciplines. One key area of interest is trajectory planning for machines in
mechatronics, computing time-dependent position, velocity, or acceleration
profiles that provide setpoints for controllers of industrial machines. In this
context, electronic cams define the repetitive motion of servo drives, commonly
defined as input point clouds and approximated by \glspl{pp}. Traditionally, the
approximation process involves solving a system of equations in closed form,
incorporating domain-specific constraints like continuity, cyclicity, or
periodicity. While computationally efficient, closed-form solutions offer
limited flexibility regarding polynomial degrees, polynomial bases, or the
integration of additional constraints.

As optimization goals grow more complex, numerical methods such as gradient
descent become beneficial. Since gradient-based optimization lies at the heart
of training (deep) neural networks, modern \gls{ml} frameworks like TensorFlow
or PyTorch come with a range of recent optimizers. In previous work \cite{WH24}
\cite{HW23}, we utilize these optimizers directly to fit $\mathcal{C}^k$-continuous
\glspl{pp} for the use in the electronic cam domain.

In this work we extend the problem setting to a constrained multi-objective
optimization problem by adding an energy term as a domain-specific loss term
and investigate (i) convergence and (ii) the pareto front under this extension.
More precisely, given input points $x_1 \le \dots \le x_n \in \nR$ and
respective target values $y_1, \dots, y_n \in \nR$, we look for a \gls{pp}
function $f$ that minimizes both the approximation loss $\ell_2$ and energy
loss $\ell_E = \int_I f''(x)^2 \dif x$ over some domain $I$ that captures the
total curvature, under the constraint that $f \in \mathcal{C}^k$, where
$\mathcal{C}^k$ denotes the set of $k$-times continuously differentiable
functions $\nR \to \nR$. We form a loss $\ell_\CK$ that captures the
\enquote{amount of violation} of $\mathcal{C}^k$-continuity, which will lead to
the optimization constraint $\ell_\CK = 0$, as it is of practical relevance to 
feed servo drives with \glspl{pp} of continuity class $\mathcal{C}^k$ to 
prevent excessive forces, vibrations and wear on machine parts.

Note that $\ell_E$ is related to energy in multiple ways: In a mechanical
setting, $\ell_E$ is the elastic strain energy of a rod bent as the function
graph of $f$. When $f$ is used as a cam and we approximately assume that the
acceleration $f''$ is proportional to the motor coil current then $\ell_E$ is
proportional to the copper losses in the motor coils.


%


\subsection{Prior and Related Work}

Related work in the cam approximation domain mostly either relies on closed-form
solutions, like \cite{MA2004}, or utilizes parametric functions for approximation
of given input data, like B-Spline or NURBS curves, as in \cite{NKHC2019}. The
first is limited with respect to complexity and number of optimization goals and
the latter cannot directly be utilized in industrial servo drives utilizing
non-parametric \gls{pp} functions. Our current work represents a natural
progression from the principles introduced in \cite{WH24} and \cite{HW23}.

\section{The optimization problem}
\label{sec:theproblem}

\paragraph{Parameterized model.}
Let us consider $n$ input points $x_1 \le \dots \le x_n \in \nR$ with respective
target values $y_1, \dots, y_n \in \nR$. We split the input domain $I = [x_1,
x_n]$ into $m$ sub-intervals $I_i = [\xi_{i-1}, \xi_i]$, with $1 \le i \le m$,
where $x_1 = \xi_0 \le \dots \le \xi_m = x_n$. A \gls{pp} function $f$ of
degree $d$ is defined by $m$ polynomial functions $p_1, \dots, p_m$ of degree
$d$ as
\begin{align}
  \label{eq:polynomial_model}
  p_i(x) = \sum_{j=0}^d \alpha_{i,j} x^j,
\end{align}
where $p_i$ covers the sub-domain $I_i$ of $I$.
As a parameterized model, $f$ possesses $(d+1) \cdot m$ model parameters
$\alpha_{i,j}$, which are trained via gradient descent optimizers according to
the loss functions described below.


\paragraph{Loss functions.}
We build on \cite{HW23} for the approximation loss $\ell_2$ and the
$\mathcal{C}^k$-continuity loss $\ell_\CK$. The former is defined as
\begin{align}
  \label{eq:l2_loss}
  \ell_2 = \frac{1}{n} \sum_{i=1}^n |f(x_i) - y_i|^2
\end{align}
and the latter introduces a term $\Delta_{i,j}$ that measures
the discontinuity of the $j$-th derivative at $\xi_i$ to then define
\begin{align}
  \label{eq:ck_loss}
  \ell_\CK = \frac{1}{m - 1} \sum_{i=1}^{m-1} \sum_{j=0}^k \Delta_{i,j}^2
  \quad \text{with} \quad
  \Delta_{i,j} = p^{(j)}_{i+1}(\xi_i) - p^{(j)}_i(\xi_i).
\end{align}
Recall that the $\xi_i$ are the boundary points of the domains $I_i$ of $p_i$.
Note that achieving $\ell_\CK = 0$ means that $f$ is $\mathcal{C}^k$-continuous
and $\ell_2 = 0$ means zero approximation error.

We can account for cyclicity (e.g., of the cam profile) by replacing in
\cref{eq:ck_loss} the two occurrences of $m-1$ by $m$ and generalizing
$\Delta_{i,j}$ to
\begin{align}
  \label{eq:cyclicity}
  \Delta_{i,j} = p^{(j)}_{1 + (i \bmod m)}(\xi_{i \bmod m}) -
  p^{(j)}_i(\xi_i).
\end{align}
(We may exclude the case $j=0$ when $i=m$ if we do not require periodicity, see
\cite{HW23} for details.)

\paragraph{Energy loss.}
In this work, we motivated the energy loss $\ell_E$ for $f$ as
\begin{align}
    \label{eq:loss-e-def}
    \ell_\mathrm{E} = \int_I f''(x)^2 \; \dif x
      &= \sum_{i=1}^m \int_{I_i} p''_i(x)^2 \; \dif x
\end{align}
in the introduction and further note that
\begin{align}
    \label{eq:loss-e-segment}
    \int_{I_i} p''_i(x)^2 \; \dif x = \int_{I_i} \left( \sum_{j=2}^d \alpha_{i,j} \cdot j (j-1) x^{j-2} \right)^2  \; \dif x.
\end{align}
Solving the definitive integral in \cref{eq:loss-e-segment} and plugging it
into \cref{eq:loss-e-def} gives us the energy loss in closed form as a function
of the trainable model parameters:
\begin{align}
    \ell_\mathrm{E} = \sum_{i=1}^m \sum_{j=2}^d \sum_{k=2}^d \alpha_{i,j} \alpha_{i,k} \frac{j k (j-1) (k-1)}{j+k-3}
        \cdot \left(\xi_i^{j+k-3}-\xi_{i-1}^{j+k-3}\right).
\end{align}

\paragraph{Multi-objective optimization.}

Various applications in engineering require $\mathcal{C}^k$-continuity up to
some $k$, i.e., continuous velocities or accelerations of cam profiles. Hence,
we treat $\ell_\CK = 0$ as a constraint in the 2-objective optimization problem
\begin{mini}
  {}{\beta \ell_2 + (1-\beta)\ell_\mathrm{E}}{\label{eq:constrainedproblem}}{}
  \addConstraint{\ell_\CK}{=0}{}
\end{mini}
over the trainable model parameters $\alpha_{i,j}$ and with $0 \le \beta \le
1$. We generalize \cite{HW23} by turning the problem into an unconstrained
3-objective optimization problem
\begin{mini}
  {}{\alpha \ell_{\text{CK}} + \beta \ell_2 + (1-(\alpha+\beta)) \ell_{\mathrm{E}}}{\label{eq:unconstrainedproblem}}{}
\end{mini}
with $\alpha, \beta \geq 0$ and $\alpha + \beta \leq 1$. Note that we can
therefore apply the unconstrained optimizers shipped with \gls{ml} frameworks
like TensorFlow. The optimization result is then processed by the algorithm
CKMIN from \cite{WH24} to strictly establish $\ell_\CK = 0$, which effectively
adds correction polynomials to obtain $\mathcal{C}^k$-continuity.

\section{Experimental Results and Discussion}

The code and results discussed in this section along with further experiments
and plots are provided in the public repository \cite{WH25}.

\begin{figure}[h]
  \centering
  \includegraphics[width=0.9\textwidth]{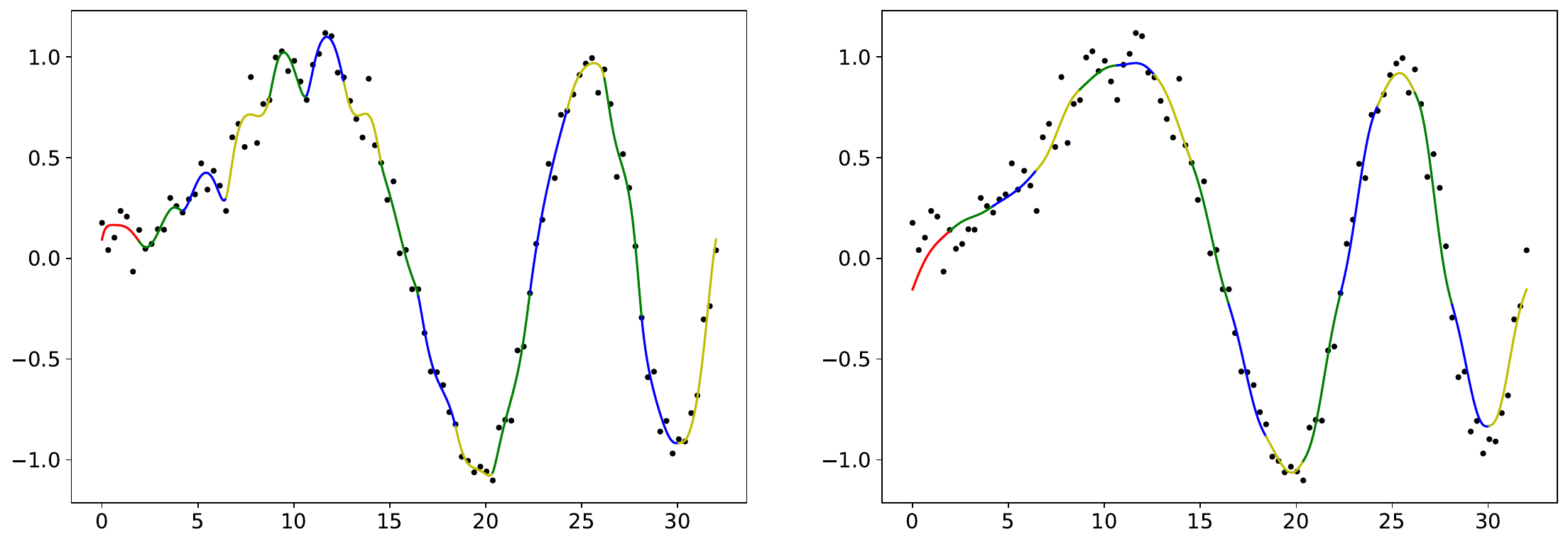}
  \caption{Results for the first two rows of \cref{tab:experimental-result}. Left: $\beta=0.9$, right: $\beta=0.45$.}
  \label{fig:curves}
 \end{figure}

For our experiment, we consider a dataset with $n=100$ input points $x_i$ 
with target values $y_i = \sin(4 \pi x_i^2 ) + n_i$, where $n_i$ is a
normal noise with zero mean and standard deviation $0.1$, produced by
\emph{numpy.random.normal}. In practice, such data could stem from
measurements. In \cref{fig:curves} the dataset is illustrated and we observe
that solely optimizing for approximation loss leads to significant curvatures
(left), while adding $\ell_\mathrm{E}$ regularizes the result (right).

\begin{figure}
  \centering
  \begin{minipage}[b]{0.45\textwidth}
    \centering
    \includegraphics[width=\linewidth]{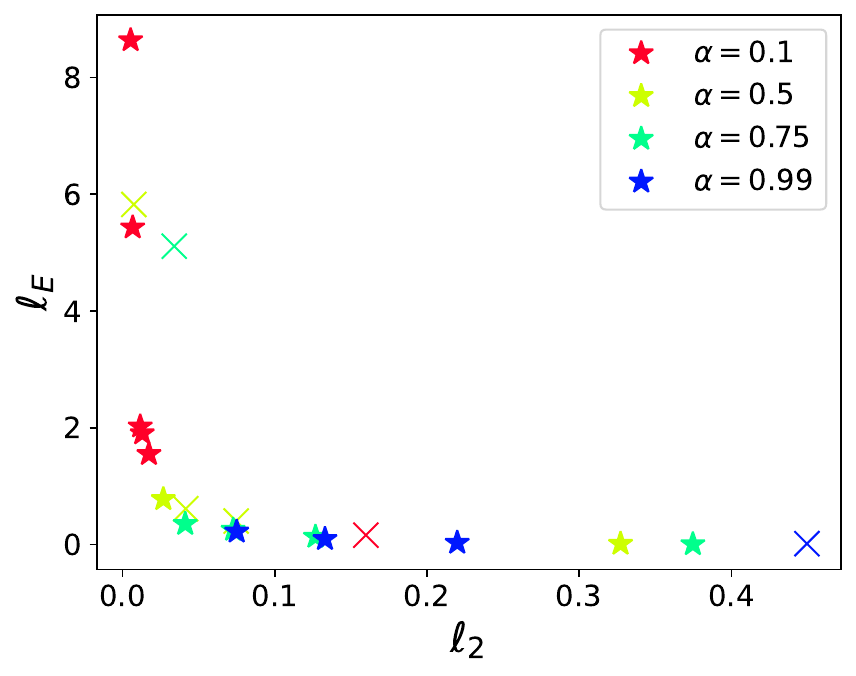}
    \caption{$\ell_\mathrm{E}$ over $\ell_2$ for different $\alpha$ and $\beta
    \in (0, 1-\alpha]$. $\star$ is Pareto optimal.}
    \label{fig:pareto-front}
  \end{minipage}
  \hspace{1em}
  \begin{minipage}[b]{0.45\textwidth}
    \centering
    \begin{tabular}{lcccr}
      \toprule
      Parameters & $\ell_2$ & \textbf{$\ell_\mathrm{E}$}\\
      \midrule
      \rowcolor[gray]{0.92}
      $\alpha=0.10, \beta=0.900$ & $0.005$ & $8.682$\\
      \rowcolor[gray]{0.92}
      $\alpha=0.10, \beta=0.450$ & $0.014$ & $1.970$\\
      $\alpha=0.50, \beta=0.500$ & $0.007$ & $5.468$\\
      $\alpha=0.50, \beta=0.250$ & $0.046$ & $0.616$\\
      $\alpha=0.75, \beta=0.250$ & $0.007$ & $5.852$\\
      $\alpha=0.75, \beta=0.125$ & $0.077$ & $0.282$\\
      $\alpha=0.99, \beta=0.010$ & $0.043$ & $4.977$\\
      $\alpha=0.99, \beta=0.005$ & $0.135$ & $0.121$\\
      \bottomrule
    \end{tabular}
    \vspace{1em}
    \captionof{table}{Results for different $\alpha, \beta$.}
    \label{tab:experimental-result}
  \end{minipage}
\end{figure}

We utilize the AMSGrad optimizer in TensorFlow to fit our \gls{pp}
model as explained in \cref{sec:theproblem}. Within our training loop, we run
$1000$ epochs with early stopping and patience of $100$ epochs to fit a
$\mathcal{C}^2$-continuous, periodic \gls{pp} of degree $7$ with $16$ polynomial
segments. \Cref{tab:experimental-result} and \cref{fig:pareto-front} show the
results after establishing $\ell_\CK=0$ using algorithm CKMIN from
\cite{WH24} for different $\alpha$ and $\beta$.
\Cref{tab:experimental-result} confirms that additionally optimizing
$\ell_\mathrm{E}$ works: Reducing $\beta$ from $0.9$ to $0.45$, shown in the first
two rows of \cref{tab:experimental-result}, reduces $\ell_\mathrm{E}$ by a
factor of $4.40$.


\section{Conclusion}

Our results demonstrate that an extension of the framework introduced in
\cite{WH24} and \cite{HW23} towards energy optimization is feasible. Results
converge without additional regularization measures and indeed improve in the
case of input data that is unfavorable for sole approximation and continuity
optimization. 
Optimizing $\ell_\mathrm{E}$ further has positive side effects: \Cref{fig:curves}
shows that oscillations in the curve's shape are reduced, which is favorable
from a mechanical perspective.

While orthogonal bases, like Chebyshev polynomials, have shown much better
performance on $\ell_2$, preliminary tests indicated convergence is impaired
when adding $\ell_\mathrm{E}$. We leave the investigation of mitigation measures
to future work.

\begin{credits}
\subsubsection{\ackname} This work was supported by the Christian Doppler Research Association
(JRC ISIA) and the European Interreg Österreich-Bayern project BA0100172
AI4GREEN. We want to thank the reviewers for their valuable feedback.

\subsubsection{\discintname}
The authors have no competing interests to declare that are
relevant to the content of this article. 
\end{credits}
%
%
%
\bibliographystyle{splncs04}
\bibliography{literature}

\end{document}